# Decision Theoretic Foundations of Graphical Model Selection

Paola Sebastiani
Department of Actuarial Science and Statistics
City University

Marco Ramoni
Knowledge Media Institute
The Open University

## Abstract

This paper describes a decision theoretic formulation of learning the graphical structure of a Bayesian Belief Network from data. This framework subsumes the standard Bayesian approach of choosing the model with the largest posterior probability as the solution of a decision problem with a 0-1 loss function and allows the use of more general loss functions able to trade-off the complexity of the selected model and the error of choosing an over-simplified model. A new class of loss functions, called disintegrable, is introduced, to allow the decision problem to match the decomposability of the graphical model. With this class of loss functions, the optimal solution to the decision problem can be found using an efficient bottom-up search strategy.

## 1 INTRODUCTION

A Bayesian Belief Network (BBN) is defined by a a joint probability distribution over a directed acyclic graph (DAG), where nodes represent stochastic variables and arcs identify dependencies between a set of *parent* variables and a *child* variables. The independence assumptions embedded in the graph factorize the joint probability distribution into a set of conditional distributions, so that reasoning tasks can be efficiently performed. Although in their original formulation, both the graphical structure and the conditional probability distributions were supposed to be provided by domain experts, for the last ten years learning BBNs from data has been an active field of research. There are now several techniques to extract the graphical model of a BBN from data (Cooper and Herskovitz, 1992; Heckerman, 1997; Lauritzen, 1996; Whittaker, 1990) and BBNs are becoming an important tool in several machine learning and data mining applications. Among statistical techniques, Bayesian methods have the advantage of coupling expert knowledge on the domain of application with the sample information in the learning process. The standard Bayesian approach to model selection involves three distinct operations:

1. A set of possible models is identified, with their prior probabilities representing the expert belief in the ability of the models to capture the association among the variables.

2. A random sample of cases is collected, which is used to update the prior probabilities of each model into posterior probabilities, by using Bayes' Theorem.

3. The model with the largest posterior probability is selected.

The rationale behind this strategy is that the model with the largest posterior probability is the most likely on the light of the sample information. It is evident that model selection involves a decision process and therefore decision theory can be used to provide a normative foundation for it (Berger, 1985; Savage, 1972). Since the decision to be made concerns the statistical problem of selecting a model on the basis of its prior probability and information conveyed by data, the decision problem is usually referred to as a *statistical decision problem* (Berger, 1985). The decision theoretic formulation of the model selection process subsumes the standard Bayesian strategy of selecting the model with the largest posterior probability as the solution of a decision problem with a 0-1 loss function. Furthermore, it allows the use of different loss functions able to trade-off the complexity of the selected model and the error of choosing an over-simplified model, thus taking into account features of the extracted model that are important for the subsequent use made of it.

Although, in principle, the formulation and solution of this decision problem seems to be immediate, we are faced with the problem that, as the number of variables increases, a complete enumeration of all models is not feasible, and the formulation of the loss function can be too difficult. The complexity of the search in the model space is also a problem for the standard Bayesian strategy, which is typically overcome (Cooper and Herskovitz, 1992) by reducing the model selection process to a greedy search over a subset of models which are consistent with some order among the variables. We show that this strategy can be formulated as a sequential decision problem and we introduce a new class of loss functions called disintegrable that decompose the sequential decision problem into smaller independent problems which admit, as optimal, an efficient one-step-look-ahead strategy.

Next section formulates the selection of the DAG of a BBN as a decision problem, and it shows how the standard Bayesian approach to model selection is equivalent to the solution of a statistical decision problem with a 0-1 loss function. Section 3 describes the sequential decision approach to greedy model search, and its solution when the loss function is disintegrable is given in Section 4.

## 2 NORMATIVE MODEL SELECTION

In order to introduce the decision theoretic approach to model selection, we begin by considering a simple discrimination problem between two DAGs and then we will generalize the results to an arbitrary number of models.

### 2.1 MODEL DISCRIMINATION

Suppose we have two categorical variables $X_1$ and $X_2$, and a random sample $\mathcal{D}$ of $n$ cases. The task is to discriminate between two DAGs: $M_0$ specifies that $X_1$ and $X_2$ are independent variables, $M_1$ specifies that $X_2$ is a parent variable of $X_1$. The standard Bayesian solution to this problem is to assign prior probabilities $p(M_0)$ and $p(M_1)$, use the available data to compute the posterior probabilities $p(M_0|\mathcal{D})$ and $p(M_1|\mathcal{D})$ and then choose the model with the largest posterior probability. Given that:

$$p(M_i|\mathcal{D}) = \frac{p(M_i, \mathcal{D})}{p(\mathcal{D})} = \frac{p(M_i)p(\mathcal{D}|M_i)}{p(\mathcal{D})}$$

where $p(\mathcal{D})$ is the marginal probability of the data, and $p(\mathcal{D}|M_i)$ is the marginal likelihood, the model selection is based on the value of the ratio

$$r = \frac{p(M_0)p(\mathcal{D}|M_0)}{p(M_1)p(\mathcal{D}|M_1)},$$

from which the following decision rule is derived: if $r < 1$, $M_1$ is chosen, if $r > 1$, $M_0$ is chosen, and if $r = 1$ then the two models are equivalent. When $p(M_0) = p(M_1)$, $r$ is the Bayes factor, i.e. $r = p(\mathcal{D}|M_0)/p(\mathcal{D}|M_1)$.

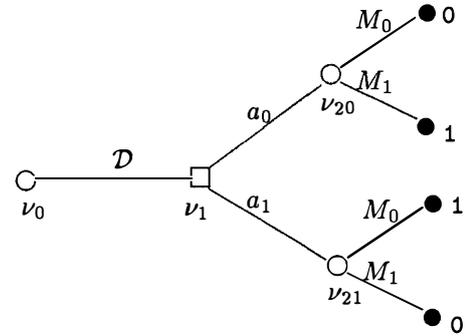

Figure 1: Decision tree for the decision problem with 0-1 loss function

Within this formulation, the discrimination between $M_0$ and $M_1$ reveals to be a *statistical decision problem* in which the *true* state of Nature is an element of the set $\mathcal{M} = \{M_0, M_1\}$, the *action space* $\mathcal{A}$ is the set $\{a_0, a_1\}$, where $a_i$ is the action "choose $M_i$", data is the sample $\mathcal{D}$, and the *loss function* is the 0-1 function defined for $(M, a) \in \mathcal{M} \times \mathcal{A}$ as:

$$L(M, a) = \begin{array}{|c|cc|} \hline & a_0 & a_1 \\ \hline M_0 & 0 & 1 \\ M_1 & 1 & 0 \\ \hline \end{array}.$$

The decision problem is represented by the decision tree in Figure 1, in which circles represent random nodes, squares represent decision nodes, and leaves (black circles) are value nodes. Thus, we first collect data $\mathcal{D}$, then use the data to choose either action $a_0$ or $a_1$. The loss incurred if the true state of Nature reveals to be $M_i$ and the action chosen is $a_j$ is then represented in the leaf nodes, and it is 1 if $i \neq j$ and 0 otherwise. The optimal decision, i.e. the *Bayesian action*, is found by minimizing the expected loss. This is done by "averaging out" and "folding back" (Raiffa and Schlaifer, 1961). From the terminal nodes, we compute the expected loss at random nodes, given everything on the left of the node, and we minimize the expected loss at the decision nodes. The expected loss of the decision $a_i$, also known as the *risk* of the decision $a_i$, at the node $\nu_{2i}$, is

$$R(a_i, \mathcal{D}) = E\{L(M, a_i)|\mathcal{D}\} = \begin{cases} p(M_0|\mathcal{D}) & i = 1 \\ p(M_1|\mathcal{D}) & i = 0 \end{cases}$$

where the expectation is over the conditional distribution of $M$ given $\mathcal{D}$. The Bayesian action is found at node $\nu_2$ by choosing:

$$a^* = \arg\min_i\{R(a_i, \mathcal{D})\}$$

which is equivalent to the decision rule $r$ found above. The risk of the Bayesian action is called the *Bayesian risk* and it is the posterior probability of the model chosen.

This formalization has the advantage that we can generalize the decision problem by using different loss functions, without modifying the prior probabilities of $M_0$ and $M_1$. The 0-1 loss function penalizes the choice of an unnecessary complex model ($M_1$ instead of $M_0$) as the choice of an over-simplified model ($M_0$ instead of $M_1$). In general, we may wish to penalize the two errors in different ways, and this can be done by using the 0-L loss:

$$L(M, a) = \begin{array}{c|cc} & a_0 & a_1 \\ \hline M_0 & 0 & l_{01} \\ M_1 & l_{10} & 0 \end{array}.$$

Thus, $l_{ij}$ is the loss incurred if the state of Nature is represented by model $M_i$ and model $M_j$ is chosen. In this case, the risks at $\nu_{20}$ and $\nu_{21}$ are

$$R(a_i, \mathcal{D}) = \begin{cases} p(M_0|\mathcal{D})l_{01} & i = 1 \\ p(M_1|\mathcal{D})l_{10} & i = 0 \end{cases}$$

and, therefore, the Bayesian action is $a_1$ if $p(M_1|\mathcal{D}) > l_{01}p(M_0|\mathcal{D})/l_{10}$ or, equivalently, if $p(\mathcal{D}|M_1) > (l_{01}p(M_0)p(\mathcal{D}|M_0))/(l_{10}p(M_1))$.

**Example 1** Let $X_1$ be a discrete variable with $c_1$ states, and let $X_2$ be a discrete variable with $c_2$ states. For simplicity we will denote the events $X_2 = x_{2j}$ and $X_1 = x_{1k}$ by $x_{2j}$ and $x_{1k}$. Model $M_0$ specifies that the two variables are independent and, conditional on $M_0$, we can parameterize $p(x_{2j}|\theta^{(0)}) = \theta_j$ and $p(x_{1k}|\theta^{(0)}) = \theta_{.k}$. Thus, $\theta^{(0)}$ is the parameter vector associated to $M_0$. Model $M_1$ specifies that $X_2$ is a parent of $X_1$, and the associated parameter vector $\theta^{(1)}$ has elements $\theta_j = p(x_{2j}|\theta^{(1)})$ and $p(x_{1k}|x_{2j}, \theta^{(1)}) = \theta_{jk}$. It is well known (see for instance the recent review by Heckerman (1997)) that the marginal likelihood $p(\mathcal{D}|M_i)$ is easily found under the assumptions that: 1. *the sample is complete;* 2. *the cases are independent, given the parameter vector $\theta^{(i)}$ associated to $M_i$;* 3. *the prior distribution of the parameters is a Dirichlet distribution;* 4. *the parameters are marginally independent.*

Suppose that, given $M_0$, $\theta_J \equiv (\theta_1, \ldots, \theta_{c_2}) \sim D(\alpha/c_2, \ldots, \alpha/c_2)$ and $\theta_{.K} \equiv (\theta_{.1}, \ldots, \theta_{.c_1}) \sim D(\alpha/c_1, \ldots, \alpha/c_1)$, where $D(\alpha_1, \ldots, \alpha_n)$ is a Dirichlet distribution with hyper-parameters $(\alpha_1, \ldots, \alpha_n)$. Given $M_1$, we assume that $\theta_{jK} \equiv (\theta_{j1}, \ldots, \theta_{jc_1}) \sim$ $D(\alpha/(c_1c_2), \ldots, \alpha/(c_1c_2))$. These parameterizations ensure that, a priori, the probabilities $p(x_{2j})$, $p(x_{1k})$ and $p(x_{1k}|x_{2j})$ are all uniform and are based on the same total prior precision on $\theta^{(i)}$. Let $n(x_{1k}|x_{2j})$ be the sample frequency of $(x_{1k}|x_{2j})$, so that $n(x_{2j}) = \sum_{k=1}^{c_1} n(x_{1k}|x_{2j})$ is the sample frequency of $x_{2j}$, and $n(x_{1k}) = \sum_{j=1}^{c_2} n(x_{1k}|x_{2j})$ is the sample frequency of $x_{1k}$. Then:

$$p(\mathcal{D}|M_0) = \prod_{j=1}^{c_2} \frac{\Gamma(\alpha)\Gamma(\alpha/c_2 + n(x_{2j}))}{\Gamma(\alpha+n)\Gamma(\alpha/c_2)}$$
$$\times \prod_{k=1}^{c_1} \frac{\Gamma(\alpha)\Gamma(\alpha/c_1 + n(x_{1k}))}{\Gamma(\alpha+n)\Gamma(\alpha/c_1)};$$
$$p(\mathcal{D}|M_1) = \prod_{j=1}^{c_2} \frac{\Gamma(\alpha)\Gamma(\alpha/c_2 + n(x_{2j}))}{\Gamma(\alpha+n)\Gamma(\alpha/c_2)}$$
$$\times \prod_{k=1}^{c_1} \frac{\Gamma(\alpha/c_2)\Gamma(\alpha/(c_1c_2) + n(x_{1k}|x_{2j}))}{\Gamma(\alpha/c_2 + n(x_{2j}))\Gamma(\alpha/(c_1c_2))}$$

and the Bayesian action under a general 0-L loss function is to choose model $M_1$ if

$$\frac{\prod_{j,k} \frac{\Gamma(\alpha/c_2)\Gamma(\alpha/(c_1c_2) + n(x_{1k}|x_{2j}))}{\Gamma(\alpha/c_2 + n(x_{2j}))\Gamma(\alpha/(c_1c_2))}}{\prod_k \frac{\Gamma(\alpha)\Gamma(\alpha/c_1 + n(x_{1k}))}{\Gamma(\alpha+n)\Gamma(\alpha/c_1)}} > \frac{l_{01}p(M_0)}{l_{10}p(M_1)}.$$

If the effect of the prior hyper-parameters is negligible, for instance when the frequencies $n(x_{1k}|x_{2j})$ are large, and $p(M_0) = p(M_1)$, then $r$ is equivalent to the likelihood ratio test (Berger, 1985), and the Bayesian rule becomes equivalent to the classical significance test. In this case, $M_0$ is accepted if $2\log r < \chi^2_{\alpha,(c_1-1)(c_2-1)}$, where $\chi^2_{\alpha,(c_1-1)(c_2-1)}$ is the $(1-\alpha)\%$ quantile of a $\chi^2$ distribution on $(c_1-1)(c_2-1)$ degrees of freedom. Thus, when $2(\log l_{10}p(M_1) - \log l_{01}p(M_0)) = \chi^2_{\alpha,(c_1-1)(c_2-1)}$ the decision rules in both approaches are identical. Note that, in the classical approach, the region of the sample space in which $M_1$ is rejected as true model, is a function of the number of states of the variables $X_1$ and $X_2$. □

## 2.2 GENERAL SOLUTION

The framework described in the previous section can be generalized to the situation in which we have a set of variables $\mathcal{X} = \{X_1, \cdots, X_I\}$, and we look for a DAG to represent the independence assumptions among the variables. Let $\mathcal{M} = \{M_0, M_1, \ldots, M_g\}$ be the set of all possible models, representing the possible states of Nature. We will keep the symbol $M_0$ to denote the *null* model: the model of mutual independence among the variables in $\mathcal{X}$. These $g+1$ models determine the action space which is now $\mathcal{A} = \{a_0, a_1, \ldots, a_g\}$, and the action $a_i$ represents the choice of model $M_i$. The loss function is given by a $(g+1) \times (g+1)$ table:

$$L(M,a) = \begin{array}{c|cccc} & a_0 & a_1 & \ldots & a_g \\ \hline M_0 & 0 & l_{01} & \ldots & l_{0g} \\ M_1 & l_{10} & 0 & \ldots & l_{1g} \\ \vdots & \vdots & \vdots & & \vdots \\ M_g & l_{g0} & l_{g1} & \ldots & 0 \end{array} \quad (1)$$

where $l_{ij}$ is the loss incurred if the true state of Nature is $M_i$, and $M_j$ is chosen. The larger number of possible models induces an expansion of the decision tree in Figure 1. Node $\nu_1$ will have $g+1$ branches, each of them corresponding to one of the possible actions. Each branch corresponding to the action $a_j$ will terminate in a random node $\nu_{2j}$ corresponding to the "revelation" of the true state of Nature and it will then be expanded into $g+1$ branches representing the possible states of Nature. Thus, at the leaves of each branch there will be the loss incurred: $l_{ij}$, $i = 0, \ldots, g$. The Bayesian action $a^*$ is then found by minimizing the expected loss. The risk of the action $a_j$ at node $\nu_{2j}$ is $R(a_j, \mathcal{D}) = \sum_{i=0}^{g} l_{ij} p(M_i|\mathcal{D})$ and $a^* = \arg\min_j\{R(a_j, \mathcal{D})\}$. With a 0-1 loss function, the Bayesian action corresponds to the standard Bayesian solution.

**Theorem 1** *For the decision problem with data $\mathcal{D}$, state of Nature $\mathcal{M} = \{M_0, M_1, \ldots, M_g\}$, action space $\mathcal{A} = \{a_0, a_1, \ldots, a_g\}$, where $a_i$ is the choice of $M_i$, and loss function defined as in (1), with $l_{ij} = 1$ for $i \neq j$, the Bayesian action is $a_i$ if $p(M_i|\mathcal{D}) > p(M_j|\mathcal{D})$ for all $j \neq i$.*

*Proof.* It is enough to show that $R(a_i, \mathcal{D}) - R(a_j, \mathcal{D}) < 0$ for all $j \neq i$. Suppose that $p(M_i|\mathcal{D}) > p(M_j|\mathcal{D})$ for all $j \neq i$, then $R(a_i, \mathcal{D}) - R(a_j, \mathcal{D}) = p(M_j|\mathcal{D}) - p(M_i|\mathcal{D}) < 0$ for all $j \neq i$. □

With a generic loss function, however, the Bayesian action may not be so simple to identify.

**Example 2** Let $\mathcal{X} = \{X_1, X_2, X_3\}$ and denote by $c_i$ the number of states of $X_i$. Suppose that $X_2, X_3$ are known to be marginally independent, and that they can be both parents of $X_1$, but $X_1$ cannot be parent of $X_2, X_3$. The set of possible models to be considered is limited to $\mathcal{M} = \{M_0, M_2, M_3, M_{23}\}$ which are given in Figure 2. Thus, the action space is given by the four possible actions of choosing one of the four models. Suppose we use the loss function $L(M, a) =$

| | $a_0$ | $a_3$ | $a_2$ | $a_{23}$ |
|---|---|---|---|---|
| $M_0$ | 0 | $kc_3$ | $kc_2$ | $k(c_3+c_2)$ |
| $M_3$ | $h$ | 0 | $k(c_3+c_2)$ | $kc_3$ |
| $M_2$ | $h$ | $k(c_3+c_2)$ | 0 | $kc_2$ |
| $M_{23}$ | $2h$ | $h$ | $h$ | 0 |

(2)

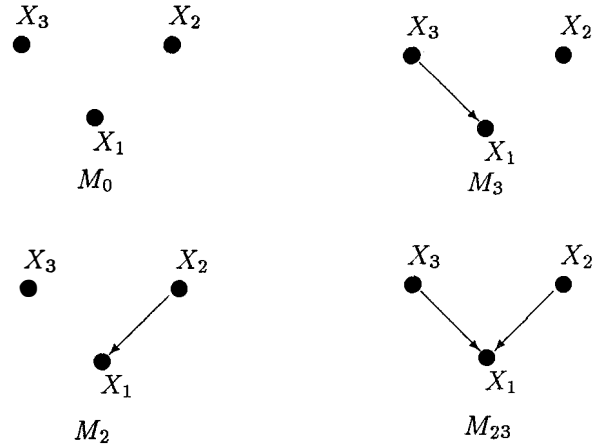

Figure 2: Models in Example 2.

where $h$ and $k$ are positive constants. Thus, the loss for choosing an unnecessarily complex model is an increasing function of the number of states of the variables. On the other hand, the loss for the choice of an over-simplified model is an increasing function of the number of possible parents left. The four risks are:

$$R(a_0, \mathcal{D}) = h\{p(M_3|D) + p(M_2|D)\} + 2hp(M_{23}|D)$$
$$R(a_3, \mathcal{D}) = kc_3 p(M_0|D) + k(c_3 + c_2)p(M_2|D) + hp(M_{23}|D)$$
$$R(a_2, \mathcal{D}) = kc_2 p(M_0|D) + k(c_3 + c_2)p(M_3|D) + hp(M_{23}|D)$$
$$R(a_{23}, \mathcal{D}) = k(c_3 + c_2)p(M_0|D) + kc_3 p(M_3|D) + kc_2 p(M_2|D)$$

and the Bayesian action is the one with minimum risk. Suppose, for instance, that $p(M_3|D) = p(M_2|D) = p$, $p(M_{23}|D) = 2p$ and $p(M_0|D) = 1 - 4p$, with $p < 0.25$. Suppose further $c_3 = 2$ and $c_2 = 3$. Then,

$$\begin{array}{rcl} R(a_0, \mathcal{D}) & = & 6hp \\ R(a_3, \mathcal{D}) & = & 2k - 3kp + 2hp \\ R(a_2, \mathcal{D}) & = & 3k - 7kp + 2hp \\ R(a_{23}, \mathcal{D}) & = & 5k - 15kp. \end{array}$$

We have $R(a_3, \mathcal{D}) < R(a_2, \mathcal{D})$, so that $M_3$ is preferred to $M_2$, although the two models have the same posterior probability. A 0-1 loss function would not allow us to discriminate between $M_2$ and $M_3$. Note also that $R(a_3, \mathcal{D}) = 2k - 3kp + 2hp \leq 5k - 15kp = R(a_{23}, \mathcal{D})$ if $p \leq 3k/(12k + 2h)$. Since $3k/(12k + 2h) < 0.25$, then $R(a_3, \mathcal{D}) < R(a_{23}, \mathcal{D})$ if $p < 3k/(12k + 2h)$. Furthermore, we have the following inequalities:

$$R(a_0, \mathcal{D}) < R(a_3, \mathcal{D}) \quad \text{iff} \quad p < \frac{2k}{4h + 3k}$$
$$R(a_0, \mathcal{D}) < R(a_{23}, \mathcal{D}) \quad \text{iff} \quad p < \frac{5k}{6h + 15k}$$

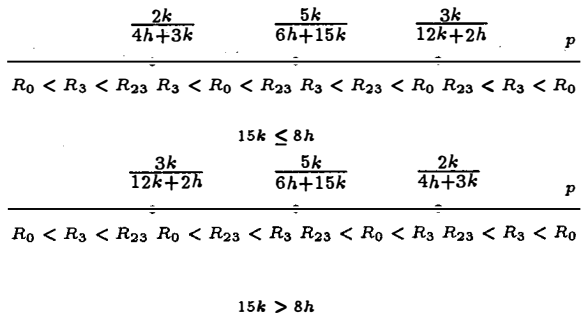

Figure 3: Ordering among risks for Example 2 when $0 < p < 0.25$: $R_0 \equiv R(a_0, \mathcal{D})$; $R_3 \equiv R(a_3, \mathcal{D})$; $R_{23} \equiv R(a_{23}, \mathcal{D})$.

$$\frac{2k}{4h+3k} \leq \frac{5k}{6h+15k} \quad \text{iff} \quad 15k \leq 8h.$$

The ordering among risks given in Figure 3 induces the following decision rule:

If $15k \leq 8h$ then

$$a^* = \begin{cases} a_0 & \text{if } 0 < p \leq 2k/(4h+3k) \\ a_3 & \text{if } 2k/(4h+3k) < p \leq 3k/(12k+2h) \\ a_{23} & \text{if } 3k/(12k+2h) < p \leq 0.25 \end{cases}$$

If $15k > 8h$, then

$$a^* = \begin{cases} a_0 & \text{if } 0 < p \leq 5k/(6h+15k) \\ a_{23} & \text{if } 5k/(6h+15k) < p \leq 0.25 \end{cases}$$

Thus, the standard Bayesian solution of choosing model $M_{23}$, is replaced by a more complex strategy, in which model $M_{23}$ is chosen if its probability is larger than $3k/(6k+h)$. In other word, a complex model is chosen when there is enough evidence in favor of it.

A simpler loss function would yield a simpler decision rule. Suppose, for instance, that we decide to penalize the choice of a complex model uniformly, via the loss function:

$$L(M,a) = \begin{array}{c|cccc} & a_0 & a_3 & a_2 & a_{23} \\ \hline M_0 & 0 & 1 & 1 & 1 \\ M_3 & h & 0 & h & h \\ M_2 & h & h & 0 & h \\ M_{23} & 2h & 2h & 2h & 0 \end{array} \quad (3)$$

The risks of the four actions are:

$R(a_0, \mathcal{D}) = h\{p(M_3|D) + p(M_2|D)\} + 2hp(M_{23}|D)$
$R(a_3, \mathcal{D}) = p(M_0|D) + hp(M_2|D) + 2hp(M_{23}|D)$
$R(a_2, \mathcal{D}) = p(M_0|D) + hp(M_3|D) + 2hp(M_{23}|D)$
$R(a_{23}, \mathcal{D}) = p(M_0|D) + hp(M_3|D) + hp(M_2|D).$

For instance, $a_0$ is the Bayesian action if $p_0 > hp_3$, $p_0 > hp_2$ and $p_0 > 2hp_{23}$: the null model is chosen if its posterior probability is $h$-times larger than the posterior probabilities of the two models with one arc only, and twice as large as the posterior probability of the model with two arcs. In doing so, we let the choice of the model depend on the complexity of the network to be chosen, and we favor the choice of more complex models. Note that the comparison between models $M_3$ and $M_2$ depends only on their posterior probabilities, and it is therefore consistent with this strategy, since both models have the same number of arcs. □

Clearly, as the number of variables increases, so does the complexity of the decision problem to solve, and we are faced with two problems:

(1) The definition of the loss function becomes too complex.

(2) The number of possible models explodes.

Problem (2) has been examined by several authors, and a solution is to reduce the model selection process to a greedy search over a subset of models which are consistent with some order among the variables, by taking advantage of the multiplicative form of the posterior probability of a model. We can similarly decompose the decision problem into sub-problems to match the decomposition of the model search.

## 3 DECOMPOSABLE DECISION PROBLEMS

Suppose we have an order on the variables in $\mathcal{X} = \{X_1, ..., X_I\}$, so that $X_i \prec X_j$ if $X_i$ cannot be parent of $X_j$. Let $\mathcal{P}_i = \{X_{i1}, ..., X_{iq_i}\}$ be the set of possible parents of $X_i$. Thus, $\mathcal{P}_i$ is the empty set if $X_i$ is a root node. Consider a DAG $M$, that specifies, for each node variable $X_i$, the set of its parents, and let them be $\Pi_i$. Denote by $n(x_{ik}|\pi_{ij})$, $i = 1, ..., I, j = 1, ..., q_i, k = 1, ..., c_i$, the sample frequency of $(x_{ik}, \pi_{ij})$, so that $n(\pi_{ij}) = \sum_{k=1}^{c_i} n(x_{ik}|\pi_{ij})$ is the sample frequency of $\pi_{ij}$. We also invoke Assumptions 1 – 4 listed in the description of Example 1 and assume that, given $M$, the vector of parameters $\theta_{ij} = (\theta_{ij1}, ..., \theta_{ijc_i})$ associated to the conditional distribution of $X_i|\pi_{ij}$ has a Dirichlet distribution $D(\alpha_{ij1}, ..., \alpha_{ijc_i})$. Thus, $\alpha_{ij} = \sum_k \alpha_{ijk}$ is the prior precision of $\theta_{ij}$. It is shown by Cooper and Herskovitz (1992) that the posterior probability of $M$ is

$$p(M|\mathcal{D}) \propto p(M) \prod_{i=1}^{I} \prod_{j=1}^{q_i} \prod_{k=1}^{c_i} \frac{\Gamma(\alpha_{ij})\Gamma(\alpha_{ijk} + n(x_{ik}|\pi_{ij}))}{\Gamma(\alpha_{ij} + n(\pi_{ij}))\Gamma(\alpha_{ijk})}.$$

Note that $p(M|\mathcal{D})$ has a multiplicative structure, since $p(M|\mathcal{D})$ is given (up to a proportionality constant) by the product, over the sets $\{X_i, \mathcal{P}_i\}$, $i = 1, ..., I$, of the

probabilities

$$p(M^i|\mathcal{D}) = \prod_{j=1}^{q_i}\prod_{k=1}^{c_i} \frac{\Gamma(\alpha_{ij})\Gamma(\alpha_{ijk}+n(x_{ik}|\pi_{ij}))}{\Gamma(\alpha_{ij}+n(\pi_{ij}))\Gamma(\alpha_{ijk})}$$

associated to the local dependencies in $\{X_i, \mathcal{P}_i\}$. This property is exploited by Cooper and Herskovitz (1992) to derive a bottom-up search strategy over the sets $\{X_i, \mathcal{P}_i\}$ known as K2 algorithm. In order to capture this search strategy in a decision theoretic framework, we need to define an algebraic structure on the set of models. Let $\mathcal{M}^i$ be the set of possible models to be explored in each $\{X_i, \mathcal{P}_i\}$. This set can be represented by a matroid with $q_i + 1$ levels. Each level contains $C(q_i, j) = \binom{q_i}{j}$ models with $j$ arcs pointing to $X_i$. We shall denote one such a model by $M^i_{co(q_i,j)}$, where $co(q_i, j)$ is a possible combination of $j$ indexes out of the $q_i$ indexes $i1, i2, ..., iq_i$ that identify the variables in $\mathcal{P}_i$. The number of models to be explored in $\mathcal{M}^i$ is then $2^{q_i} = \sum_{j=1}^{q_i} \binom{q_i}{j}$. Let $M^i_0, M^i_1, ..., M^i_{q_i}$ be elements of $\mathcal{M}^i$ where $M^i_0$ is the null model, and each $M^i_j$ is the model with only one arc from $X_{ij}$ pointing to $X_i$, We can regard the set $\mathcal{M}^i$ as generated by $M^i_0, M^i_1, ..., M^i_{q_i}$ via the sum of models $\uplus$ which is defined as follows. Let $M^i_{co(q_i,j)}$ and $M^i_{co(q_i,l)}$ be elements of $\mathcal{M}_i$, then $M^i_{co(q_i,j)} \uplus M^i_{co(q_i,l)} = M^i_{co(q_i,m)}$ is the model containing all arcs pointing to $X_i$, that are specified by the two models. This algebraic structure decomposes every model with more than one arc pointing to $X_i$, into the sum of models with one arc only, e.g. $M^i_{i1,...,ik} = M^i_0 \uplus M^i_1 \uplus ... \uplus M^1_k$. For instance, in the four models in Figure 2, $M_{23}$ is the sum $M_2 \uplus M_3$. Furthermore, if $M^i_{co(q_i,j)}$ and $M^j_{co(q_j,l)}$ are models in $\mathcal{M}^i$ and $\mathcal{M}^j$, we define $M^i_{co(q_i,j)} \uplus M^j_{co(q_j,l)}$ as the model containing all arcs specified by the two models. In this way, every DAG for the variables in $\mathcal{X}$ can be regarded as a sum of models in $\mathcal{M}^i$, $i = 1, ..., I$.

The decision problem describing the search over the sets $\mathcal{M}^i$ is now a pseudo sequential statistical decision problem: we use the term "pseudo" because we do not have a sequential collection of data. A typical branch of the decision tree is represented in Figure 4. Once data are collected at node $\nu_0$, at node $\nu_1$ we choose an action from the action space $\mathcal{A}^1 = \{a^1_0, a^1_1, ..., a^1_{2^{q_1}}\}$, corresponding to all possible models $\mathcal{M}^1$. The consequence of each action is represented by the possible models in $\mathcal{M}^1$. Next, we have the decision node $\nu_2$ with action space $\mathcal{A}^2 = \{a^2_0, a^2_1, ..., a^2_{2^{q_2}}\}$, corresponding to a choice in the set $\mathcal{M}^2$. The consequence of each action is represented by the possible models in $\mathcal{M}^2$, and so on. The decision problem terminates after $I$ steps, corresponding to the $I$ sets $\mathcal{M}^i$. It is evident that we can regard each action space $\mathcal{A}^i$ as generated by $a^i_0, a^i_1, ..., a^i_{q_i}$, with $a^i_j$ defined as choosing model $M^i_j$. The choice of a model with more than one arc is then the sum of the generating actions, i.e. $a^i_{jl} = a^i_j \uplus a^i_l$ is "choose the models with arcs from $X_{ij}$ and from $X_{il}$" and so on. The terminal nodes in the decision tree report the loss incurred when the true state of Nature correspond to the sum of models "revealed" along the branch, and the action chosen is the sum of actions taken at the $I$ decision nodes in the branch. As in Section 2, the problem is solved by averaging out and folding back. Thus, we start from the terminal nodes in the tree, we find the action that minimizes the expected loss given everything is on the left, and then we proceed backward, by finding optimal actions and folding back the tree. The decision nodes are replaced by value nodes reporting the Bayesian risks of the optimal actions. Next section will show that, in our case, there exists a class of loss function which admit a solution easy to find.

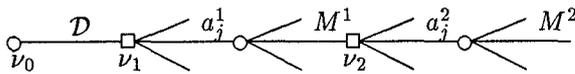

Figure 4: A typical branch of the decision tree describing the sequential decision problem.

## 4 DISINTEGRABLE LOSS FUNCTIONS

The algebraic structure on the set of all possible models translates into a similar structure on the loss function. Consider first the local decision problem in $\mathcal{M}^i$. The loss function can be built up from simple loss functions associated to the $q_i$ pair-wise comparisons between $M^i_0$ and each $M^i_j$ as follows. Let $L^i_j(M^i, a^i)$ be the 0-L loss function for discriminating between models $M^i_0$ and $M^i_j$, were $a^i_j$ is the action "choose model $M^i_j$". Thus, $L^i_j(M^i, a^i)$ is defined over the space $\{M^i_0, M^i_j\} \times \{a^i_0, a^i_j\}$ as:

$$L^i_j(M^i, a^i) = \begin{array}{|c|c|c|} \hline & a^i_0 & a^i_j \\ \hline M^i_0 & 0 & l_{0j} \\ \hline M^i_j & l_{j0} & 0 \\ \hline \end{array}.$$

We define the sum $\oplus$ of $L^i_j(M^i, a^i)$ and $L^i_l(M^i, a^i)$ as the loss function defined over the set $[\{M^i_0, M^i_j\} \uplus$

$\{M_0^i, M_l^i\}] \times [\{a_0^i, a_j^i\} \uplus \{a_l^i, a_{jl}^i\}]$ by:

$$L_{jl}^i(M,a) = L_j^i(M^i,a^i) \oplus L_l^i(M^i,a^i)$$

|   | $a_0^i$ | | | | $a_l^i$ | | | |
|---|---|---|---|---|---|---|---|---|
| $M_0^i$ | 0 + | | $a_0^i$ | $a_j^i$ | $l_{0l}$ + | | $a_0^i$ | $a_j^i$ |
|         |     | $M_0^i$ | 0 | $l_{0j}$ |           | $M_0^i$ | 0 | $l_{0j}$ |
|         |     | $M_j^i$ | $l_{j0}$ | 0 |         | $M_j^i$ | $l_{j0}$ | 0 |
| $M_l^i$ | $l_{l0}$ + | | $a_0^i$ | $a_j^i$ | 0 + | | $a_0^i$ | $a_j^i$ |
|         |     | $M_0^i$ | 0 | $l_{0j}$ |           | $M_0^i$ | 0 | $l_{0j}$ |
|         |     | $M_j^i$ | $l_{j0}$ | 0 |         | $M_j^i$ | $l_{j0}$ | 0 |

=

|   | $a_0^i$ | $a_j^i$ | $a_l^i$ | $a_{jl}^i$ |
|---|---|---|---|---|
| $M_0^i$ | 0 | $l_{0j}$ | $l_{0l}$ | $l_{0j} + l_{0l}$ |
| $M_j^i$ | $l_{j0}$ | 0 | $l_{0l} + l_{j0}$ | $l_{0l}$ |
| $M_l^i$ | $l_{l0}$ | $l_{l0} + l_{0j}$ | 0 | $l_{0j}$ |
| $M_{jl}^i$ | $l_{j0} + l_{l0}$ | $l_{l0}$ | $l_{j0}$ | 0 |

By iteratively computing the sum of all $q_i$ loss functions, we derive the loss function for the local decision problem in $\mathcal{M}^i$ which is defined on $\mathcal{M}^i \times \mathcal{A}^i$:

$$L^i(M,a) = L_{i1}^i(M^i,a^i) \oplus L_{i2}^i(M^i,a^i) \oplus ... \oplus L_{iq_i}^i(M^i,a^i).$$

We will call a loss function which can be obtained in such a way a *locally disintegrable* loss function.

The rationale behind the choice of this loss function is that the error in choosing $M_{co(q_i,j)}^i$ instead of $M_{co(q_j,l)}^j$ is in the number of arc differences between the two models and we penalize this error by summing up the losses corresponding to each arc difference. Consider, for instance, the four models in Figure 2. The error in choosing either models $M_3$ or $M_2$ compared to $M_0$ is only in one arc. If $M_2$ is chosen instead of $M_3$, the error is given by adding the arc from $X_2$ to $X_1$ (i.e. choosing $M_2$ instead of $M_0$) and removing the arc from $X_3$ to $X_1$ (i.e. choosing $M_0$ instead of $M_3$.) We thus penalize this error by summing up the two losses corresponding to choosing $M_2$ instead of $M_0$, and to choosing $M_0$ instead of $M_3$.

Let now $L^i(M,a)$, $i = 1,...,I$, be the disintegrable loss functions associated to the $I$ local decision problems. We define as *globally disintegrable* for the sequential decision problem the loss function generated as:

$$L(\{M^1 \uplus M^2 \uplus ... \uplus M^I\}, \{a^1 \uplus a^2 \uplus ... \uplus a^I\})$$
$$= L^1(M^1,a^1) \oplus L^2(M^2,a^2) \oplus ... \oplus L^I(M^I,a^I).$$

Thus, $L(M,a)$ is a table of dimensions $\prod_i 2^{q_i} \times \prod_i 2^{q_i}$. Each row represent a possible state of Nature $M$ given by the sum of models in each $\mathcal{M}^i$, which are themselves sum of generating models. The columns represent the possible actions computed as sum of the actions chosen in each $\mathcal{A}^i$. Given the additive structure of $L(M,a)$, it is easily seen that the loss assigned to each terminal node of the decision tree is the loss cumulated along a branch. Consider one of the terminal decision nodes. The action space is $\mathcal{A}^I = \{a_0^I, a_1^I, ..., a_{q_I}^I\}$, the true state of Nature is one of the models in $\mathcal{M}^I$. The risk of the decision $a_j^I$ is then

$$R^I(a_j^I, \mathcal{D}, M_{h1}^1,...,M_{h(I-1)}^{I-1}, \{a_{k1}^1,...,a_{k(I-1)}^{I-1}\})$$
$$= \sum_i l_{ij} p(M_i^I | \mathcal{D}, M_{h1}^1,...,M_{h(I-1)}^{I-1})$$
$$= \sum_i l_{ij} p(M_i^I | \mathcal{D})$$

where $M_{h1}^1,...,M_{h(I-1)}^{I-1}$ represents the sequence of states of Nature along the branch, $\{a_{k1}^1,...,a_{k(I-1)}^{I-1}\}$ is the sequence of actions that preceed $a_j^I$, and $l_{ij} = l_{I-1} + l_{ij}^I$, with $l_{I-1}$, representing the cumulative loss along the branch up to node $\nu_I$. Thus, the minimum risk can be found independently of $l_{I-1}$ and the Bayesian action $a^{I*}$ turns out to be the same in each of the terminal decision nodes. Hence, at each decision node $\nu_I$ we attach the Bayesian risk:

$$R^I(a^{I*}, \mathcal{D}, M_{h1}^1,...,M_{h(I-1)}^{I-1}, \{a_{k1}^1,...,a_{k(I-1)}^{I-1}\}).$$

Similar simplifications apply when we move backward to the decision node $\nu_{I-1}$. The risk of the decision $a_{k(I-1)}^{I-1}$ is:

$$R^{I-1}(a_{k(I-1)}^{I-1}, \mathcal{D}, M_{h1}^1,...,M_{h(I-2)}^{I-2}, \{a_{k1}^1,...,a_{k(I-2)}^{I-2}\})$$
$$= \sum_h R^I(a^{I*}, \mathcal{D}, M_{h1}^1,...,M_h^{I-1}, \{a_{k1}^1,...,a_{k(I-1)}^{I-1}\})$$
$$\quad \times p(M_h^{I-1} | \mathcal{D}, M_{h1}^1,...,M_{h(I-2)}^{I-2})$$
$$= \sum_h R^I(a^{I*}, \mathcal{D}, M_{h1}^1,...,M_h^{I-1}, \{a_{k1}^1,...,a_{k(I-1)}^{I-1}\})$$
$$\quad \times p(M_h^{I-1} | \mathcal{D})$$

where $M_{h1}^1,...,M_{h(I-2)}^{I-2}$ is the sequence of states of Nature along the branch up to $\nu_{I-1}$, $M^{I*}$ is the model chosen before, and

$$R^I(a^{I*}, \mathcal{D}, M_{h1}^1,...,M_h^{I-1}, \{a_{k1}^1,...,a_{k(I-1)}^{I-1}\})$$

($h = 1,...,2^{q_{I-1}}$) are the Bayesian risks attached at the $2^{q_{I-1}}$ value nodes that represent the loss incurred if the true state of Nature is one of the models $\{M_{h1}^1 \uplus ... \uplus M_{h(I-2)}^{I-2}\} \uplus M_h^{I-1}$. Given the additive nature inherited by the risk at the previous step, again the Bayesian action can be found independently of the loss cumulated until node $\nu_{I-2}$.

We then have that (1) the global disintegrability of the loss function and (2) the factorization of the joint posterior probability of a BBN decompose the sequential decision problem into local decision problems in each structure $\mathcal{M}^i$, and decisions made relative to models in $\mathcal{M}^i$ are *irrelevant* for decisions made about $\mathcal{M}^j$, for $i \neq j$. Thus, in this case, the one-step-look-ahead strategy (Berger, 1985) is optimal. We can now take advantage of the local disintegrability of the loss function $L^i$ to guide the local search in $\mathcal{M}^i$. For simplicity, we focus on $\mathcal{M}^1$, and we drop the superscript 1. The loss function is the $2^{q_1} \times 2^{q_1}$ table defined as:

$$L(M, a) = L_1(M, a) \oplus L_2(M, a) \oplus ... \oplus L_{q_1}(M, a).$$

By definition, this table has only $1 + q_1$ independent columns which correspond to the generating actions $a_0, a_1, \ldots, a_{q_1}$. Let $R_0 = R(a_0, \mathcal{D}), R_1 = R(a_1, \mathcal{D}), \ldots, R_{q_1} = R(a_{q_1}, \mathcal{D})$ be the corresponding risks. From these values, all the pair-wise comparisons can be easily generated, so that they can be performed in time linear with respect to the number of possible parents, as shown in the following example.

**Example 3** Consider the decision problem in Example 2. Define the loss functions:

$$L_3(M, a) = \begin{array}{|c|cc|} \hline & a_0 & a_3 \\ \hline M_0 & 0 & l_{03} \\ M_3 & l_{30} & 0 \\ \hline \end{array}$$

and

$$L_2(M, a) = \begin{array}{|c|cc|} \hline & a_0 & a_2 \\ \hline M_0 & 0 & l_{02} \\ M_2 & l_{20} & 0 \\ \hline \end{array}$$

Then the loss function for the decision problem is

$L(M, a) = L_3(M, a) \oplus L_2(M, a) =$

| $M$ | $a_0$ | $a_3$ | $a_2$ | $a_{23}$ |
|---|---|---|---|---|
| $M_0$ | 0 | $l_{03}$ | $l_{02}$ | $l_{03} + l_{02}$ |
| $M_3$ | $l_{30}$ | 0 | $l_{30} + l_{02}$ | $l_{02}$ |
| $M_2$ | $l_{20}$ | $l_{03} + l_{20}$ | 0 | $l_{03}$ |
| $M_{23}$ | $l_{30} + l_{20}$ | $l_{20}$ | $l_{30}$ | 0 |

where the column corresponding to action $a_{23}$ is a linear combination of the first three columns. We have the following relations among comparisons of risks:

$$R_0 - R_3 = R_2 - R_{23} \quad (4)$$
$$R_0 - R_2 = R_3 - R_{23} \quad (5)$$
$$R_0 - R_{23} = (R_0 - R_3) + (R_0 - R_2) \quad (6)$$
$$R_3 - R_2 = (R_0 - R_2) - (R_0 - R_3) \quad (7)$$

and the Bayesian action can be found by simply evaluating $R_0$, $R_3$ and $R_2$:

1. If $R_0 - R_3 < 0$ and $R_0 - R_2 < 0$, then $a_0 = a^*$, since from (6) $R_0 - R_{23} < 0$;

2. If $R_0 - R_3 > 0$ and $R_0 - R_2 < 0$, then $a_3 = a^*$, since from (5) $R_3 - R_{23} < 0$;

3. If $R_0 - R_3 < 0$ and $R_0 - R_2 > 0$, then $a_2 = a^*$, since from (4) $R_2 - R_{23} < 0$;

4. If $R_0 - R_3 > 0$ and $R_0 - R_2 > 0$, then $a_{23} = a^*$, since from (4) $R_2 - R_{23} < 0$ and from (5) $R_3 - R_{23} < 0$.

□

This result can be easily extended to any number of parents $q_i$, so that from the $q_i$ independent comparisons $R_0 - R_i$, all others can be found.

## 5 CONCLUSIONS

The effort of providing a decision theoretic foundation for the model selection process is extremely rewarding: it puts theory and methods of model selection on a firmer, normative ground and provides a better understanding of the meaning of the results achieved so far. This paper shows that the decision theoretic formulation of the model selection process generalizes the standard Bayesian strategy and allows the use of different loss functions able to trade-off the complexity of the selected model and the error of choosing an oversimplified model, thus taking into account features of the extracted model that are relevant to its use.

## References

Berger, J. (1985). *Statistical Decision Theory and Bayesian Analysis* (2nd edition). Springer-Verlag: New York. First Ed 1980.

Cooper, G., and Herskovitz, E. (1992). A Bayesian method for the induction of probabilistic networks from data. *Machine Learning*, 9, 309–347.

Heckerman, D. (1997). Bayesian networks for data mining. *Data Mining and Knowledge Discovery*, 1, 79–119.

Lauritzen, S. (1996). *Graphical Models*. Clarendon Press, Oxford.

Raiffa, H., and Schlaifer, R. (1961). *Applied Statistical Decision Theory*. MIT Press, Cambridge, Mass.

Savage, L. (1972). *The Foundations of Statistics* (2nd Revised edition). Dover, New York, NY.

Whittaker, J. (1990). *Graphical Models in Applied Multivariate Statistics*. Wiley, New York, NY.